\documentclass[conference]{IEEEtran}
\IEEEoverridecommandlockouts
\usepackage{cite}
\usepackage{amsmath,amssymb,amsfonts}
\usepackage{algorithmic}
\usepackage{graphicx}
\usepackage{subcaption} 
\usepackage{textcomp}
\usepackage{tikz,hyperref}
\usepackage{booktabs}
\usepackage{array}
\usepackage{multirow}
\newcolumntype{L}[1]{>{\centering\arraybackslash}m{#1}}
\newcolumntype{C}[1]{>{\centering\arraybackslash}p{#1}}

\usepackage{xcolor}
\def\BibTeX{{\rm B\kern-.05em{\sc i\kern-.025em b}\kern-.08em
    T\kern-.1667em\lower.7ex\hbox{E}\kern-.125emX}}

\definecolor{lime}{HTML}{A6CE39}
\DeclareRobustCommand{\orcidicon}{%
	\begin{tikzpicture}
	\draw[lime, fill=lime] (0,0) 
	circle [radius=0.16] 
	node[white] {{\fontfamily{qag}\selectfont \tiny ID}};
	\draw[white, fill=white] (-0.0625,0.095) 
	circle [radius=0.007];
	\end{tikzpicture}
	\hspace{-2mm}
}

\foreach \x in {A, ..., Z}{%
	\expandafter\xdef\csname orcid\x\endcsname{\noexpand\href{https://orcid.org/\csname orcidauthor\x\endcsname}{\noexpand\orcidicon}}
}

\begin{document}

\title{Interpretable Fuzzy Systems\\  For Forward Osmosis Desalination}

\author{\IEEEauthorblockN{Qusai Khaled{\orcidA{}}}
\IEEEauthorblockA{\textit{Jheronimus Academy of Data Science} \\
\textit{Eindhoven University of Technology}\\
Eindhoven, The Netherlands \\
qusai.khaled@ieee.org}
\and
\IEEEauthorblockN{Uzay Kaymak{\orcidB{}}}
\IEEEauthorblockA{\textit{Jheronimus Academy of Data Science} \\
\textit{Eindhoven University of Technology}\\
Eindhoven, The Netherlands \\
u.kaymak@ieee.org}
\and
\IEEEauthorblockN{Laura Genga}
\IEEEauthorblockA{\textit{School of Industrial Engineering} \\
\textit{Eindhoven University of Technology}\\
Eindhoven, The Netherlands \\
l.genga@tue.nl}

}
\maketitle

\begin{abstract}
Preserving interpretability in fuzzy rule-based systems (FRBS) is vital for water treatment, where decisions impact public health. While structural interpretability has been addressed using multi-objective algorithms, semantic interpretability often suffers due to fuzzy sets with low distinguishability. We propose a human-in-the-loop approach for developing interpretable FRBS to predict forward osmosis desalination productivity. Our method integrates expert-driven grid partitioning for distinguishable membership functions, domain-guided feature engineering to reduce redundancy, and rule pruning based on firing strength. This approach achieved comparable predictive performance to cluster-based FRBS while maintaining semantic interpretability and meeting structural complexity constraints, providing an explainable solution for water treatment applications. 

\end{abstract}


\section{Introduction}
Explainable Artificial Intelligence (XAI) has become essential for the practical implementation of machine learning in a trustworthy manner, particularly in high-risk domains where decisions in critical infrastructure directly affect humans or the environment \cite{arrieta2020explainable}. In water treatment plants, operational processes are highly non-linear and complex, making predictive modeling crucial for forecasting performance, detecting faults, and optimization. The importance of interpretability in these tasks cannot be overstated, as decisions based on these models can have significant real-world impacts. Fuzzy Rule-Based Systems (FRBs) have emerged as powerful tools in this context, balancing performance with interpretability while effectively handling non-linearity inherent in water treatment processes.

Linguistic terms in fuzzy logic facilitate design of interpretable models, maintaining a balance between accuracy and interpretability. FRBs leverage human reasoning and can be constructed using expert knowledge and data-driven methods \cite{lu2024fuzzy, varshney2023literature}. Expert-driven models define rules based on domain knowledge, while data-driven methods use machine learning techniques to build relationships from historical data \cite{ross2005fuzzy}. This synergy has led to development of interpretability taxonomies and metrics to ensure design of interpretable FRBs \cite{gacto2011interpretability, moral2021explainable}. However, balancing semantic interpretability—preserving the meaning of membership functions— with model complexity and performance is a persistent challenge, especially for skewed datasets.
Semantic interpretability gains high importance in water treatment applications where predictive models target human experts who require clear and distinct membership functions instead of brute force optimization of precise fuzzy models that shift their understanding of linguistic fuzzy sets beyond comprehension. Common solutions often rely on computationally expensive multi-objective evolutionary algorithms (MOEAs) \cite{moral2021explainable, gacto2011interpretability}. While grid partitioning (GP) offers a semantically interpretable approach to fuzzy modelling, it struggles with dimensionality and low performance \cite{guillaume2001designing}.

Despite the critical importance of balancing semantic interpretability, dimensionality, and accuracy, research addressing these trade-offs remains limited. This paper presents a human-in-the-loop approach with the objective of creating a semantically interpretable model adhering to complexity constraints and delivers performance comparable to automated methods. Using expert intuition, we integrate domain-oriented feature engineering (DFE), fixed expert-defined partitioning (FGP), rule inactivity checking (IA), and regularized global consequent estimation (GCE). The proposed DFE-FGP-IA-GCE routine is applied to predict forward osmosis desalination using a real skewed experimental dataset. Results indicate performance comparable to fuzzy clustering-based methods in terms of mean absolute error (MAE), lower modeling complexity, in terms of the number of parameters, and higher semantic interpretability as measured by distinguishability.

The remainder of this paper is organized as follows: Section II explores interpretability taxonomies and related work on the semantic interpretability of fuzzy models, particularly Takagi-Sugeno models. Section III details the design methodology for inherently interpretable FRBs. Sections IV and V present a case study that predicts the performance of water treatment, followed by a comparison with cluster-based methods. Section VI concludes with discussion and future work.
Code is publicly available at 
github.com/QusaiKhaled/DFE-FGP-IA-GCE
\section{Related Work}
\begin{figure}[b]
    \centering
    \includegraphics[width=0.7\linewidth]{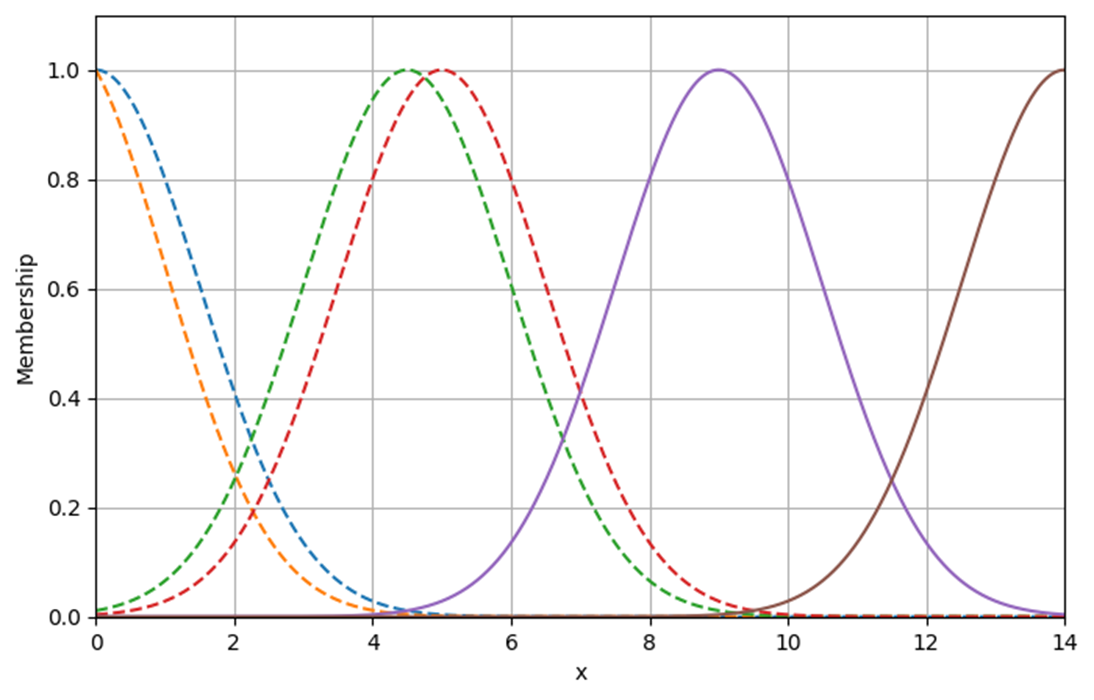}
    \caption{\small Example of fuzzy sets satisfying the distinguishability criteria, plotted in continuous lines, compared to sets violating distinguishability, visualized with dashed lines.}
    \label{fig:distinguishability}
\end{figure}

Researchers developed various frameworks for interpretability assessment in FRBs\cite{gacto2010integration}\cite{moral2021explainable}\cite{guillaume2001designing}\cite{zhou2008low}. 
Gacto et al. (2011) \cite{gacto2011interpretability} introduced a 2×2 taxonomy analyzing FRBS interpretability, examining rule base and fuzzy partition levels through complexity-based and semantic interpretability. The former addresses structural constraints like rule numbers and membership functions, while the latter focuses on meaningful linguistic variables through properties including completeness, normalization, distinguishability and complementarity. Distinguishability, one of the most frequently analyzed semantic properties, ensures fuzzy sets remain distinct and assignable to unique linguistic terms \cite{moral2021explainable}, directly impacting a model's reasoning clarity. Figure \ref{fig:distinguishability} depicts an example of distinguishable and non-distinguishable sets.

Several methods measure distinguishability. First, similarity quantified by the Jaccard index \( J(A, B)=\frac{|A \cap B|}{|A \cup B|} \) determines overlap between fuzzy sets, where lower values indicate greater distinguishability. Second, Jin et al. (2003) introduced a Gaussian-based measure \cite{jin2000fuzzy}, computing similarity \( S \) between functions \( A \) and \( B \) as {\small \( S(A, B) = \sqrt{(\mu_1 - \mu_2)^2 + (\sigma_1 - \sigma_2)^2} \)} and distinguishability as {\small \( d(A,B) = \frac{1}{1 + S(A, B)} \)}, where lower similarity represents improved distinguishability\cite{jin2003generating}. Third, Mencar's possibility measure evaluates distinguishability through maximum overlap between normal, convex and continuous sets, calculated as {\small \( P(A, B) = \sup_{x \in U} \min(\mu_A(x), \mu_B(x)) \)}, with values approaching zero indicating higher distinguishability \cite{mencar2007distinguishability}.

Various distinguishability preservation measures exist, including similarity-based merging \cite{setnes1998similarity} and graph theory-based sets dropping \cite{fuchs2020graph}. MOEAs are also commonly used to balance interpretability and accuracy objectives \cite{shukla2012review}. Using gaussian-based similarity, \cite{jin2000fuzzy} employed genetic algorithms to reduce rule redundancy, indirectly improving distinguishability. With Mencar's possibility, researchers utilized Differential Evolution to tune membership functions \cite{maisto2012possibilistic}. Cheong et al., 2003 proposed a constrained genetic algorithm approach, maintaining distinguishability by limiting overlap between membership functions and constraining maximum membership function locations \cite{cheong2003constrained}.

Grid partitioning provides an alternative way to maintain distinguishability by dividing the input space uniformly into partitions over their corresponding universe of discourse. It assigns fuzzy sets to each feature, which represent linguistic labels shared across all rules. Typically, grid partitioning is achieved by segmenting each variable's range into a fixed number of intervals, disregarding data density and distribution \cite{guillaume2001designing}\cite{zhou2008low}. All possible fuzzy sets combinations are considered in the formulation of the rule base, which could be a drawback as it results in rule redundancy, since some rules might not be fired. This method is susceptible to the curse of dimensionality\cite{eftekhari2008extracting}. However, although it is not the preferred choice in automated fuzzy systems, it is considered highly interpretable compared to other methods such as clustering \cite{guillaume2001designing}.

Grid partitioning methods have many applications where the grid structure is built and optimized \cite{zhou2008low}. The local linear model tree (LOLIMOT) algorithm is one of the key applications of grid partitioning to create interpretable neuro-fuzzy models, where the input space is divided into hyper-rectangles and then refined iteratively \cite{nelles1996basis}\cite{nelles2000comparison}. Another example is \cite{zhang2023optimized}, where authors explored the application of the adaptive neuro-fuzzy inference system (ANFIS) for predicting thermo-physical properties of hybrid nanofluids, comparing grid partitioning (GP), fuzzy c-means (FCM), and subtractive clustering (SC). All three methods demonstrated high predictive accuracy following a whole search (WS) optimization strategy. To address the curse of dimensionality in GP, the authors restricted the number of membership functions per variable, using a maximum of three MFs across four variables. While this strategy reduced dimensionality, the study does not explore whether the tunable parameters of the membership functions may have compromised the semantic interpretability of the resulting models after optimization.

Although GP methods initially produce uniform and distinguishable sets, they struggle with the dimensionality and reduced performance compared to data-dependent methods. Efforts to address these issues often rely on automatically tunable parameters, which risk compromising the semantic interpretability of the grid, particularly in real-world scenarios with skewed datasets. Limited research exists on the preservation of semantic interpretability under these challenging conditions.

\section{Design Methodology for Interpretable FRBs}
We propose a human-in-the-loop GP approach predicting forward osmosis desalination performance from skewed experimental data. Our method integrates domain-oriented feature engineering (DFE) for dimensionality reduction, fixed expert-defined partitions (FGP) preserving semantic interpretability, inactivity checking (IA) eliminating rule redundancy and regularized global consequent estimation (GCE) for mitigating rule dominance bias. This approach balances semantic and complexity-based interpretability while achieving performance comparable to clustering-based methods across both dense and sparse input regions. This section explains the compiled DFE-FGP-IA-GCE routine. We address a non-linear regression problem using a Takagi-Sugeno model, splitting our approach into two phases: the human-in-the-loop phase involving expert involvement during DFE and FGP, and the data-driven IA-GCE phase for dimensionality reduction and optimization.

\subsection{Human in the Loop (HITL) Phase}
\noindent
\textit{Domain Feature Engineering:} The number of rules in grid partitioning is given by $n^m$, where $n$ is the number of fuzzy sets per feature, and $m$ is the number of features. By selecting fewer features, we reduce both the rule base size and the complexity. Since our approach emphasizes domain expertise in designing interpretable systems, domain-driven feature engineering is key to minimizing the number of features and, in turn, the complexity. The optimal number of features for interpretability is subjective: Gacto et al. \cite{gacto2011interpretability, mccutcheon2005novel} suggest adhering to the 7 ± 2 rule for human cognitive limits \cite{mccutcheon2005novel}, while Varshney et al. \cite{varshney2023literature} recommend no more than three antecedents, and Moral et al. \cite{moral2021explainable} warn that exceeding seven compromises interpretability. In line with Occam’s razor, we simplify using DFE to reduce $m$ to three features and select three fuzzy sets $n$ per feature, linguistically assigned low, medium, and high, respectively.

In the context of forward osmosis desalination, sensory data from the water treatment plant are used as input to formulas (\ref{eq:water_flux}) and (\ref{eq:osmotic_pressure}), widely employed in membrane science to model transport phenomena using the solution-diffusion model.

\noindent
\begin{minipage}{0.45\linewidth}
\begin{equation}
\small
J_w = A \Delta \pi
\label{eq:water_flux}
\end{equation}
\end{minipage}%
\hfill
\begin{minipage}{0.45\linewidth}
\begin{equation}
\small
\pi = i \cdot M \cdot R \cdot T
\label{eq:osmotic_pressure}
\end{equation}
\end{minipage}
Here, $J_w$ is the membrane flux $L/m^2 h$, $A$ is the membrane permeability $m^3/m^2 h $ $bar$, and $\Delta \pi$ is the osmotic pressure difference in $bar$ between the feed and draw solution sides, making it a key engineered feature due to its proportional relationship with flux $J_w$. $\pi$ is calculated using the van't Hoff factor $i$, $M$ is the molarity, $mol/m^3$, $R$ is the gas constant, $J/mol \cdot K$, and $T$ is temperature in Kelvin $K$. This transformation of raw sensory data simplifies the input space while preserving domain relevance. Further details on sensory and engineered features are presented in the next section.

\textit{Fixed Grid Partitioning:} The proposed design of FGP parameters for antecedents ensures semantic interpretability through:
1) Coverage: uniform distribution of fuzzy sets over each feature's range.  
2) Normality: each set contains at least one data point with a membership value of 1.  
3) Distinguishability: limited overlap between adjacent sets with smooth transitions between linguistic terms. Taking that into consideration, and assuming Gaussian membership functions, the partition width \(\Delta U_i\) for feature \(i\), the mean \(\mu_j\) of each fuzzy set, and the shared standard deviation \(\sigma\) for all fuzzy sets within the feature are defined as follows:  

\vspace{-3mm}
\begin{equation}
\small
\Delta U_i = \frac{U_{\text{(i)max}} - U_{\text{(i)min}}}{n - 1}
\end{equation}
\vspace{-5mm}
\begin{equation}
\small
\mu_{ij} = U_{\text{(i)min}} + (j-1) \cdot \Delta U_i, \quad 
\sigma_{ij} = \frac{\Delta U_i}{n + k}
\end{equation}
\vspace{-5mm}

\noindent
where \(U_{\text{(i)min}}\) and \(U_{\text{(i)max}}\) represent the minimum and maximum values of the feature, \(n\) is the total number of fuzzy sets, \(j = 1, 2, \ldots, n\) denotes the index of each fuzzy set, determined by its position from left to right across the feature’s range,  \(k > 0\) is a tuning factor introduced to adjust the degree of overlap between adjacent membership functions. The choice of \(k\) depends on expert judgment, guided by the desired level of distinguishability. In this paper, \(k\) is set to 1 during implementation to maintain modest overlap. However, experts can refine the initial positioning and overlap of fuzzy sets as needed to align the fuzzy sets with their linguistic interpretation.

\subsection{Data-driven phase}
\noindent
\textit{Inactivity Checking:} considering a first order Takagi-Sugeno system, the structure of the rule base can be expressed as: 
\vspace{-2mm}
\begin{align*}
\small
R_{j}: \text{IF } & \text{x}_1 \text{ is } \text{mf}_{1j} \text{ and } \text{x}_2 \text{ is } \text{mf}_{2j} \text{ and } \ldots \text{ and } \text{x}_m \text{ is } \text{mf}_{mn} \\
& \text{Then } z_{j} = w_1 \cdot \text{x}_1 + w_2 \cdot \text{x}_2 + \ldots + w_m \cdot \text{x}_m + b 
\end{align*}
\vspace{-7mm}

\noindent
IA implements rule pruning to reduce dimensional complexity in grid partitioned rule bases. Rules with firing strength below a specific threshold are removed as redundant. Using product t-norm for antecedents, firing strength $\beta$ is expressed per rule as follows:
\vspace{-3mm}

\noindent
\begin{minipage}{0.49\linewidth}
\begin{equation}
\small
\beta = mf_{{1}}(x_1) \ldots mf_{{n}}(x_n)
\label{eq:beta}
\end{equation}
\end{minipage}%
\hfill
\begin{minipage}{0.49\linewidth}
\begin{equation}
\small
mf(x) = \exp\left( -\frac{1}{2} \left( \frac{x - \mu}{\sigma} \right)^2 \right)
\label{eq:mf}
\end{equation}
\end{minipage}
\vspace{1mm}

\noindent
The firing strength for rule \( r \) is estimated by summing the firing strengths of that rule across all data points. Let \( \beta_{r} \) represent the cumulative firing strength for \( r \), and let \( R \) be the total number of rules. The normalized firing strength for rule \( r \), denoted as \( \tilde{\beta}_{r} \), is then:
\vspace{-2mm}
\begin{equation}
\small
\tilde{\beta}_r = \frac{\beta_r}{\sum_{j=1}^{R} \beta_j}.
\end{equation}
To maintain the compactness of the rule base as a property desired for both semantic and structural interpretability\cite{alonso2015interpretability}, any rule with a normalized firing strength \( \tilde{\beta}_r \) below a threshold of \( 10^{-2} \) is removed.

\textit{Global Consequent Estimation:} Next, for GCE, we consider the formula for predicting output of first-order Takagi-Sugeno system:

\vspace{-4mm}
\begin{equation}
\small
f(x|\theta) = \frac{\sum_{r=1}^{R} \left( b_r + \sum_{i=1}^{m} w_{ri} x_i \right) \prod_{j=1}^{n} \exp\left(-\frac{1}{2} \frac{(x_j - \mu_{rj})^2}{\sigma_{rj}^2}\right)}{\sum_{r=1}^{R} \prod_{j=1}^{n} \exp\left(-\frac{1}{2} \frac{(x_j - \mu_{rj})^2}{\sigma_{rj}^2}\right)} \label{8}
\end{equation}

\noindent
Here $\theta$ is the least squares estimate vector that includes the consequent parameters to be estimated. Since the antecedent parameters \(\mu\) and \(\sigma\) expressed in the Gaussian formula have been estimated in the grid partitioning stage, the consequent paramaters \(b\) and \(w\) are estimated using ordinary least squares estimates while adding $L2$ ridge regularization as a penalty term to ensure all rules contribute to the prediction and prevent overfitting.  

Consider a simple rule base with 2 rules, 2 features, and 2 fuzzy sets each. The design matrix is estimated, considering that the output is composed of the weighted average of both rules:
\vspace{-4mm}
\begin{equation}
\small
f(x|\theta) = \frac{\beta_1 \cdot z_1 + \beta_2 \cdot z_2}{\beta_1 + \beta_2} = \frac{\beta_1 \cdot z_1 }{\beta_1 + \beta_2} + \frac{\beta_2 \cdot z_2 }{\beta_1 + \beta_2}. \label{9}
\end{equation}

\noindent
Each of the two aggregated terms corresponds to a rule that is estimated as the value of rule consequent $z$ multiplied by its normalized firing strength $\beta$. The two terms can be written as:
\begin{equation}
\frac{\beta_1 \cdot (w_{11} \cdot x_1 + w_{21} \cdot x_2 + b_1) }{\beta_1 + \beta_2} + \frac{\beta_2 \cdot (w_{12} \cdot x_1 + w_{22} \cdot x_2 + b_2) }{\beta_1 + \beta_2} \label{10}
\end{equation}
Ross (2005) \cite{ross2005fuzzy} describes each of the two terms as the regression vector $\epsilon$, where $f(x|\hat{\theta}) = \hat{\theta} \mathcal{E}(x) =\hat{\theta} (\epsilon_1 + \epsilon_2$). Consequently, epsilon composes the building element of the design matrix in our approach, where $m$ corresponds to the number of rules and $n$ the number of data points,

\vspace{-3mm}
\begin{equation}
\small
\Phi = \begin{pmatrix}
\epsilon_{11} & \epsilon_{12} & \cdots & \epsilon_{1m} \\
\epsilon_{21} & \epsilon_{22} & \cdots & \epsilon_{2m} \\
\vdots & \vdots & \ddots & \vdots \\
\epsilon_{n1} & \epsilon_{n2} & \cdots & \epsilon_{nm}
\end{pmatrix}. \label{11}
\end{equation}
\vspace{-3mm}

After estimating the values of \(\Phi\), the $L2$ formula is implemented iteratively to obtain the weights and the bias of the consequents by solving for damped least squares with QR decomposition through the following objective function,

\vspace{-3mm}
\begin{equation}
\small
\hat{\mathbf{w}}_{\text{ridge}} = \arg\min_{\mathbf{w}} \left( \| \Phi \mathbf{w} - \mathbf{y} \|_2^2 + \lambda \| \mathbf{w} \|_2^2 \right), \label{12}
\end{equation}
\vspace{-5mm}

\noindent
where $\Phi \in \mathbb{R}^{n \times m}$ is the design matrix constructed from the input data, $\mathbf{y} \in \mathbb{R}^{n}$ is the observed output vector, and $\mathbf{w} \in \mathbb{R}^{m}$ represents the consequent parameters to be estimated. While the antecedent parameters are fixed, the estimated consequent parameters are regularized, the performance is unlikely to be optimal, for that case, grid search based optimization is performed to tune the consequents by searching for the optimal regularization parameter $\lambda$ corresponding to lowest mean absolute error MAE.

\section{Predicting the performance of forward osmosis desalination}
We employ our method as a regression problem for predicting the performance of water treatment process named forward osmosis (FO). Below, we provide a background on this technology and description of the dataset used in this study. Then we introduce the sequence of steps taken following the methodology. 

\subsection{Forward Osmosis Desalination}
As an emerging technology, forward osmosis has been explored in the context of desalination and water treatment, pharmaceutical industry, food processing and power generation, but its commercial applications remain limited \cite{cath2006forward}. As an alternative low-energy desalination process to reverse osmosis, it functions by taking advantage of the existing natural phenomenon of osmotic pressure. Thus, it is less energy consuming than other methods, and less prone to fouling and accumulation of unwanted substances on membrane surfaces, reducing the pretreatment requirements and making it an affordable choice \cite{linares2014forward}.

In \cite{jawad2020modeling}, lab-scale experimental dataset was compiled from several studies to develop an artificial neural network for predicting permeate flux based on nine input variables, such as water temperature and velocity. The best performance (R squared 94.5\%) was achieved with 35 neurons, 4 layers and logistic sigmoid function as an activation function. Then, it was compared to a multi-linear regression model which achieved $R^2$ values of 51.6\%, suggesting non-linear relationship among variables of the dataset. Hence, a fuzzy rule-based model would be well-suited due to its ability to handle nonlinear relationships. Table \ref{tab:dataset_description} provides a description of the features and their types, where membrane flux is the target variable the models predict and translate into the productivity of the process measured in $L/(m^{2}.h)$.
\begin{table}[t]
    \centering
    \scriptsize
    \begin{tabular}{l L{3.5cm} C{1.0cm} C{1.5cm}}
        \toprule
        \textbf{No.} & \textbf{Feature Name} & \textbf{Type} & \textbf{Units} \\
        \midrule
        1 & Membrane Type & Binary     & -      \\
        2 & Membrane Orientation & Binary & -  \\
        3 & FS Molarity & Continuous & mol/m\textsuperscript{3} \\
        4 & DS Molarity & Continuous & mol/m\textsuperscript{3} \\
        5 & DS Molecular Weight & Continuous & g/mol \\
        6 & FS Velocity & Continuous & cm/s \\
        7 & DS Velocity & Continuous & cm/s \\
        8 & FS Temperature & Continuous & Kelvin \\
        9 & DS Temperature & Continuous & Kelvin \\
        10 & Membrane Flux & Continuous & $L/(m^{2}.h)$ \\
        \bottomrule
    \end{tabular}
    \caption{\small Description of the dataset features. The full dataset contains 709 samples and 12 features. DS: draw solution, FS: feed solution.}
    \label{tab:dataset_description}
\end{table}

\begin{figure}[b]
    \centering
    \includegraphics[width=0.7\linewidth]{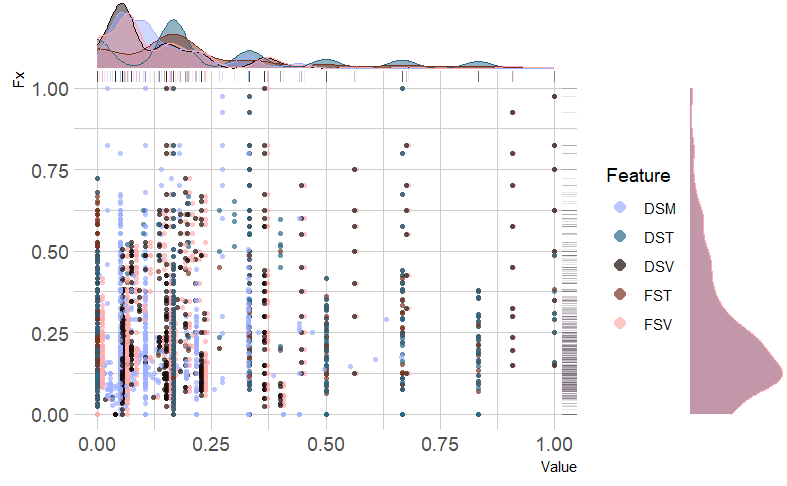}
    \caption{\small Visualization depicting the skewness of dataset features relative to the flux \( F_x \). DSM: Draw solution molarity, DST: Draw solution temperature, DSV: Draw solution velocity, FST: Feed solution temperature, FSV: Feed solution velocity.}
    \label{fig:skew}
\end{figure}

Figure \ref{fig:skew} illustrates the skewness characteristics of the dataset by visualizing five normalized input features versus corresponding membrane flux. Marginal distributions provide additional insights into feature distributions along the x-axis, while the flux distribution is represented on the y-axis. The plot highlights clustering towards lower values of the features, while emphasizing the density of data points through overlaid bar code plots aligned with feature distributions.

\subsection{Dimensionality Reduction}
 
Using formulas \ref{eq:water_flux} and \ref{eq:osmotic_pressure}, original sensory features are transformed to engineered dataset in Table \ref{tab:membrane_data}. To reduce the number of features from five to three, we split the data based on binary features for type and orientation, creating four subsets, each with three inputs and one output. A separate model is developed for each subset. Since each plant has a fixed membrane configuration, omitting these features does not affect the input-output relationship, with each model applying to one type and orientation configuration.
\begin{table}[t]
    \centering
    \scriptsize
    \begin{tabular}{l L{1.0cm} C{1.0cm} C{1.0cm} C{1.0cm} C{1.0cm}}
        \toprule
        MT & MO & DSmw (g/mol) & \(\Delta P\) (bar) & Velocity Mean (cm/s) & Membrane Flux (LMH) \\
        \midrule
        0 & 1 & 79 & 295 & 30 & 20 \\
        1 & 0 & 79 & 161 & 30 & 18 \\
        0 & 1 & 79 & 16  & 30 & 11 \\
        1 & 0 & 58 & 0   & 45 & 3  \\
        0 & 1 & 58 & 2   & 45 & 18 \\
        \bottomrule
    \end{tabular}
    \caption{\small Sample rows from the engineered dataset, MT: membrane type, MO: membrane orientation, DSmw: draw solution molecular weight, \(\Delta P\): pressure difference.}
    \label{tab:membrane_data}
\end{table}

\subsection{Rule base construction}
Pre-processing includes dataset splitting into training/testing sets and Z-score normalization to aid least squares estimation convergence. With three fuzzy sets per input variable (low, medium, high) and three input variables with one output, the initial rule base comprises 27 rules. Table \ref{tab:fuzzy_decision_matrix_conditions} summarizes these rule conditions, with each unique combination forming a distinct fuzzy system rule.

\begin{table}[b]
    \centering
    \scriptsize
    \begin{tabular}{L{1.5cm} | C{1.5cm} C{1.5cm} C{1.5cm}}
        \toprule
        \textbf{DSmw} & \multicolumn{3}{c}{\textbf{$\Delta P$}} \\
        \cmidrule{2-4}
        & $\Delta P\_mf1$ & $\Delta P\_mf2$ & $\Delta P\_mf3$ \\
        \midrule
        DSmw\_mf1 & V\_mf1, V\_mf2, V\_mf3 & V\_mf1, V\_mf2, V\_mf3 & V\_mf1, V\_mf2, V\_mf3 \\
        \midrule
        DSmw\_mf2 & V\_mf1, V\_mf2, V\_mf3 & V\_mf1, V\_mf2, V\_mf3 & V\_mf1, V\_mf2, V\_mf3 \\
        \midrule
        DSmw\_mf3 & V\_mf1, V\_mf2, V\_mf3 & V\_mf1, V\_mf2, V\_mf3 & V\_mf1, V\_mf2, V\_mf3 \\
        \midrule
        \multicolumn{4}{c}{\textbf{V}} \\
        \bottomrule
    \end{tabular}
    \caption{\small Fuzzy decision matrix for the antecedents DSmw, $\Delta P$, and V. Total fuzzy set combinations correspond to 27 rules.}
    \label{tab:fuzzy_decision_matrix_conditions}
\end{table}
\noindent
Then, in the same manner on handling rule redundancy, rule pruning is performed using the normalized firing strength of each rule obtained from the training data. Next, based on the selected threshold, omission of redundant rules have resulted in reducing the number of rules to 15.

\subsection{Estimating the consequents}
The consequent weights are then obtained using formulas (\ref{9} - \ref{12}). Considering the semantics associated with each membership function, final rule base is demonstrated in Table \ref{tab:rule_base_with_tuned}. Interestingly, the weights of the consequents are within similar ranges as the result of ridge regularization, while experimentation with Lasso regularization has produced negligible weights that would diminish the impact of some rules completely. In contrast, the use of no regularization has led to weights of extremely large values.

\begin{table*}[htbp]
\scriptsize
\centering
\begin{tabular}{|c|c|c|c|l|l|}
\hline
\textbf{Rule} & \textbf{DSmw}  & \textbf{\(\Delta P\)}  & \textbf{V}  & \textbf{Fx}                                   & \textbf{Tuned Fx} \\ \hline
1             & low            & low                    & low         & \(0.09 \cdot \text{DSmw} + 1.22 \cdot \Delta_P - 0.12 \cdot V + 2.75\) & \(0.15 \cdot \text{DSmw} + 0.56 \cdot \Delta P - 0.15 \cdot V + 0.22\) \\ \hline
2             & low            & low                    & medium      & \(1.26 \cdot \text{DSmw} + 1.51 \cdot \Delta_P + 0.49 \cdot V + 4.16\) & \(-0.08 \cdot \text{DSmw} + 0.69 \cdot \Delta P + 0.52 \cdot V + 0.27\) \\ \hline
3             & low            & low                    & high        & \(0.65 \cdot \text{DSmw} + 3.73 \cdot \Delta_P + 2.07 \cdot V - 0.83\) & \(-0.03 \cdot \text{DSmw} + 0.90 \cdot \Delta P + 0.41 \cdot V + 0.03\) \\ \hline
4             & low            & medium                 & low         & \(-0.40 \cdot \text{DSmw} + 0.86 \cdot \Delta_P - 0.10 \cdot V + 0.81\) & \(0.10 \cdot \text{DSmw} + 0.70 \cdot \Delta P + 0.01 \cdot V - 0.56\) \\ \hline
5             & low            & medium                 & medium      & \(1.03 \cdot \text{DSmw} + 1.18 \cdot \Delta_P + 1.97 \cdot V + 2.91\) & \(-0.18 \cdot \text{DSmw} + 0.38 \cdot \Delta P + 1.39 \cdot V + 0.10\) \\ \hline
6             & low            & medium                 & high        & \(0.44 \cdot \text{DSmw} + 1.85 \cdot \Delta_P + 1.69 \cdot V - 0.57\) & \(-0.26 \cdot \text{DSmw} + 0.16 \cdot \Delta P + 1.10 \cdot V + 0.34\) \\ \hline
7             & low            & high                   & low         & \(-0.86 \cdot \text{DSmw} + 0.56 \cdot \Delta_P - 0.03 \cdot V - 0.91\) & \(0.00 \cdot \text{DSmw} - 0.04 \cdot \Delta P + 0.11 \cdot V - 0.11\) \\ \hline
8             & medium         & low                    & low         & \(0.00 \cdot \text{DSmw} + 0.01 \cdot \Delta_P + 2.13 \cdot V + 2.57\) & \(-0.10 \cdot \text{DSmw} + 0.14 \cdot \Delta P + 0.47 \cdot V - 0.20\) \\ \hline
9             & medium         & low                    & medium      & \(-0.14 \cdot \text{DSmw} - 0.64 \cdot \Delta_P + 3.46 \cdot V - 2.81\) & \(-0.23 \cdot \text{DSmw} + 0.15 \cdot \Delta P - 0.06 \cdot V - 0.23\) \\ \hline
10            & medium         & medium                 & low         & \(0.36 \cdot \text{DSmw} + 0.36 \cdot \Delta_P + 8.77 \cdot V + 6.24\) & \(-0.73 \cdot \text{DSmw} + 0.09 \cdot \Delta P + 0.71 \cdot V + 0.59\) \\ \hline
11            & medium         & medium                 & medium      & \(2.26 \cdot \text{DSmw} + 1.86 \cdot \Delta_P - 3.31 \cdot V - 14.88\) & \(-0.03 \cdot \text{DSmw} - 0.29 \cdot \Delta P - 0.09 \cdot V - 0.48\) \\ \hline
12            & medium         & high                   & low         & \(-0.21 \cdot \text{DSmw} + 3.90 \cdot \Delta_P + 7.38 \cdot V - 4.17\) & \(-0.08 \cdot \text{DSmw} + 0.09 \cdot \Delta P + 0.28 \cdot V - 0.17\) \\ \hline
13            & high           & low                    & low         & \(-0.05 \cdot \text{DSmw} + 2.25 \cdot \Delta_P - 1.06 \cdot V + 3.21\) & \(-0.14 \cdot \text{DSmw} + 0.01 \cdot \Delta P + 0.00 \cdot V - 0.01\) \\ \hline
14            & high           & low                    & medium      & \(0.76 \cdot \text{DSmw} - 0.20 \cdot \Delta_P + 0.10 \cdot V + 0.35\) & \(0.04 \cdot \text{DSmw} + 0.01 \cdot \Delta P + 0.00 \cdot V + 0.00\) \\ \hline
15            & high           & medium                 & low         & \(-0.12 \cdot \text{DSmw} + 0.89 \cdot \Delta_P - 0.49 \cdot V + 1.16\) & \(0.59 \cdot \text{DSmw} + 0.28 \cdot \Delta P - 0.10 \cdot V + 0.22\) \\ \hline
\end{tabular}
\caption{\small Rule base created from grid partitioning, Tuned Fx represents weights generated after the grid search.}
\label{tab:rule_base_with_tuned}
\end{table*}

\begin{table}[htbp]
\centering
\scriptsize
\setlength{\tabcolsep}{2.5pt}
\begin{tabular}{|c|c|c|c|c|c|c|c|}
\hline
\multirow{3}{*}{\textbf{R}} & \multicolumn{6}{c|}{\textbf{Antecedents}} & \multirow{3}{*}{\textbf{Fx}} \\ \cline{2-7}
& \multicolumn{2}{c|}{\textbf{DSmw}} & \multicolumn{2}{c|}{\textbf{\(\Delta P\)}} & \multicolumn{2}{c|}{\textbf{V}}  \\ \cline{2-7}
& \(\mu\) & \(\sigma\) & \(\mu\) & \(\sigma\) & \(\mu\) & \(\sigma\)  \\ \hline
1 & 0.4 & 0.5 & 0.1 & 0.6 & -0.3 & 0.3 & \(13.5 - 2.4\text{DSmw} + 3.8\Delta_P + 3.6V\) \\ \hline
2 & 1.1 & 1.2 & 1.0 & 1.2 & -1.3 & 2.4 & \(14.2 - 2.0\text{DSmw} + 3.0\Delta_P + 7.4V\) \\ \hline
3 & 0.5 & 0.3 & -0.3 & 0.1 & -0.3 & 0.2 & \(10.5 - 1.6\text{DSmw} + 3.4\Delta_P - 2.0V\) \\ \hline
4 & 0.1 & 0.7 & 0.2 & 0.6 & -0.2 & 0.4 & \(12.9 - 2.3\text{DSmw} + 3.1\Delta_P + 3.0V\) \\ \hline
5 & 0.1 & 0.7 & 0.2 & 0.6 & -0.2 & 0.4 & \(12.9 - 2.3\text{DSmw} + 3.1\Delta_P + 3.0V\) \\ \hline
6 & 0.5 & 0.3 & -0.3 & 0.1 & -0.3 & 0.2 & \(10.5 - 1.6\text{DSmw} + 3.4\Delta_P - 2.0V\) \\ \hline
7 & 0.1 & 0.7 & 0.2 & 0.6 & -0.2 & 0.4 & \(12.9 - 2.3\text{DSmw} + 3.1\Delta_P + 3.0V\) \\ \hline
8 & 0.1 & 0.7 & 0.2 & 0.6 & -0.2 & 0.4 & \(12.9 - 2.3\text{DSmw} + 3.1\Delta_P + 3.0V\) \\ \hline
9 & 0.4 & 0.5 & 0.1 & 0.6 & -0.3 & 0.3 & \(13.5 - 2.4\text{DSmw} + 3.8\Delta_P + 3.6V\) \\ \hline
10 & 0.1 & 0.7 & 0.2 & 0.6 & -0.2 & 0.4 & \(12.9 - 2.3\text{DSmw} + 3.1\Delta_P + 3.0V\) \\ \hline
11 & 0.5 & 0.3 & -0.3 & 0.1 & -0.3 & 0.2 & \(10.5 - 1.6\text{DSmw} + 3.4\Delta_P - 2.0V\) \\ \hline
12 & 0.5 & 0.3 & -0.3 & 0.1 & -0.3 & 0.2 & \(10.5 - 1.6\text{DSmw} + 3.4\Delta_P - 2.0V\) \\ \hline
13 & 0.1 & 0.7 & 0.2 & 0.6 & -0.2 & 0.4 & \(12.9 - 2.3\text{DSmw} + 3.1\Delta_P + 3.0V\) \\ \hline
14 & 0.5 & 0.3 & -0.3 & 0.1 & -0.3 & 0.2 & \(10.5 - 1.6\text{DSmw} + 3.4\Delta_P - 2.0V\) \\ \hline
15 & 0.4 & 0.5 & 0.1 & 0.6 & -0.3 & 0.3 & \(13.5 - 2.4\text{DSmw} + 3.8\Delta_P + 3.6V\) \\ \hline
16 & 0.5 & 0.3 & -0.3 & 0.1 & -0.3 & 0.2 & \(10.5 - 1.6\text{DSmw} + 3.4\Delta_P - 2.0V\) \\ \hline
17 & 0.1 & 0.7 & 0.2 & 0.6 & -0.2 & 0.4 & \(12.9 - 2.3\text{DSmw} + 3.1\Delta_P + 3.0V\) \\ \hline
18 & 0.1 & 0.7 & 0.2 & 0.6 & -0.2 & 0.4 & \(12.9 - 2.3\text{DSmw} + 3.1\Delta_P + 3.0V\) \\ \hline
\end{tabular}
\caption{\small Rule base created from clustering-based method.}
\label{tab:rule_base_extended}
\end{table}

\subsection{Tuning the consequents}
Mean Absolute Error is chosen as a metric to assess performance against the test set, which equals 1.986 LMH. The error is considered relatively high with respect to the normalized range of the membrane flux 6.159 LMH, translating into MAPE of 32\%. Therefore, further tuning is needed to produce better performance. While maintaining fixed antecedent parameters, a grid search for the consequent parameters is done by looping around the regularization parameter $\lambda$ with the objective of obtaining parameters resulting in the least error.
100 values of $\lambda$ were tested within range \( \left[ 0.001, 10 \right] \), during which each run produced a different set of consequents weights, which was added to the model and tested for performance. The grid search has shown that the optimal model was achieved at $\lambda =1.314$ where the minimum MAE was 0.486. Based on the regularization and grid search procedure, MAPE was reduced from 32\% to 7\% percent.
The parameters of the optimized model are also included in Table \ref{tab:rule_base_with_tuned} under Tuned Fx. Notably, it was possible to achieve a substantial reduction in the error although the antecedent parameters remained fixed. Although further optimization of the antecedents could further reduce the error, doing so would compromise the semantic interpretability of the model. Specifically, adjusting the antecedents parameters could alter the pre-defined membership function boundaries based on domain expertise, potentially shifting them away from their original, interpretable positions in real-world settings.

\section{Comparison to clustering based fuzzy systems}
The proposed approach is compared with fuzzy clustering-based methods, known for outperforming grid-based methods \cite{benmouiza2019clustered}\cite{oladipo2022performance}. This comparison aims to determine if the method achieves comparable results while maintaining interpretability constraints. Performance metrics, MAE and MAPE, are used alongside semantic interpretability quantified through distinguishability metric $d$ with Gaussian-based parametric similarity, providing numerical basis for evaluating the approach's competitive advantages.

The clustering model is created from pyFUME \cite{fuchs2020pyfume}, a python package for automated fuzzy model estimation based on clustering. To maintain complexity-based interpretability, the package utilizes graph-based GRABS method for rule base simplification \cite{fuchs2020graph}. The package automatically generates a rule base compatible with Simpful python package, a python library for fuzzy logic reasoning \cite{spolaor2020simpful}. For the sake of comparison, the exact post-processed engineered data referred in Table \ref{tab:membrane_data} is used in pyFUME. Both models share the same initial number of rules, first order Takagi-Sugeno system and the same type of membership functions.
To estimate the antecedents based on clustering, Gustaffson-Kessel (GK) algorithm is selected. Clustering is performed in the input-output space. Pyfume automatically partitions the data into the selected number of clusters. Each cluster corresponds to a rule. Then consequents paramaters are estimated based on least squares estimates.

\begin{figure}[b]
    \centering
    \includegraphics[width=0.75\linewidth]{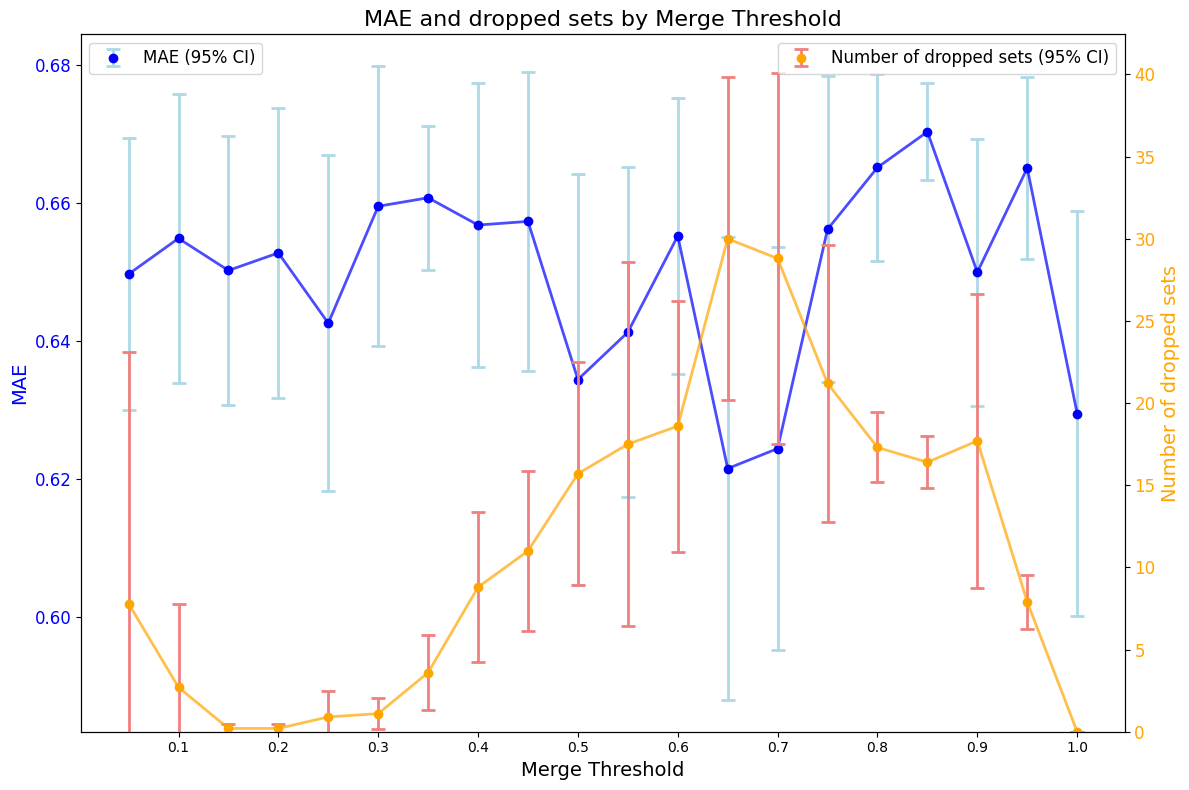}
    \caption{\small Results of GRABS simplification.}
    \label{fig:GRABS}
\end{figure}

With graph-based Simplification (GRABS) method, rather than simplifying the rules base by omitting redundant rules bearing low firing strength, similarity-based rule base simplification is used, which looks at similar fuzzy sets with high degree of overlapping, then merge them based on Jaccard similarity index \cite{fuchs2020graph}. In a similar manner to setting threshold for removing redundant rules, GRABS sets merging threshold below which merging takes place. To determine optimum threshold, a parameter sweep was conducted in range \( \left[ 0.00, 1.00 \right] \) with 0.05 steps and 10 iterations per threshold. Thresholds 0 and 1 are neglected, since 0 threshold means all sets will be merged, and 1 means no merging takes place. This process resulted in a total of 180 runs. The results are visualized in Figure \ref{fig:GRABS}. Changing the merge threshold did not affect the performance of the model significantly, but it heavily affected the number of merged (dropped sets), an optimal threshold is observed at 0.65, where average number of dropped sets is around 30.

After exploratory experiments to identify the optimal merge threshold, we conducted exploitation - akin to fine-tuning - by running the GRABS method for 100 iterations at the identified threshold of 0.65, similar to the grid search procedure used to boost grid partitioning performance. The optimal results in Table \ref{tab:rule_base_extended} show the generated rule base. Notably, the clustering-based model has significantly more parameters than the grid-partitioning model because each cluster has unique parameters, while the grid model shares parameters across all rules, as shown in Table \ref{tab:rule_base_with_tuned}.
\begin{figure}[b]
    \centering
    \includegraphics[width=0.7\linewidth]{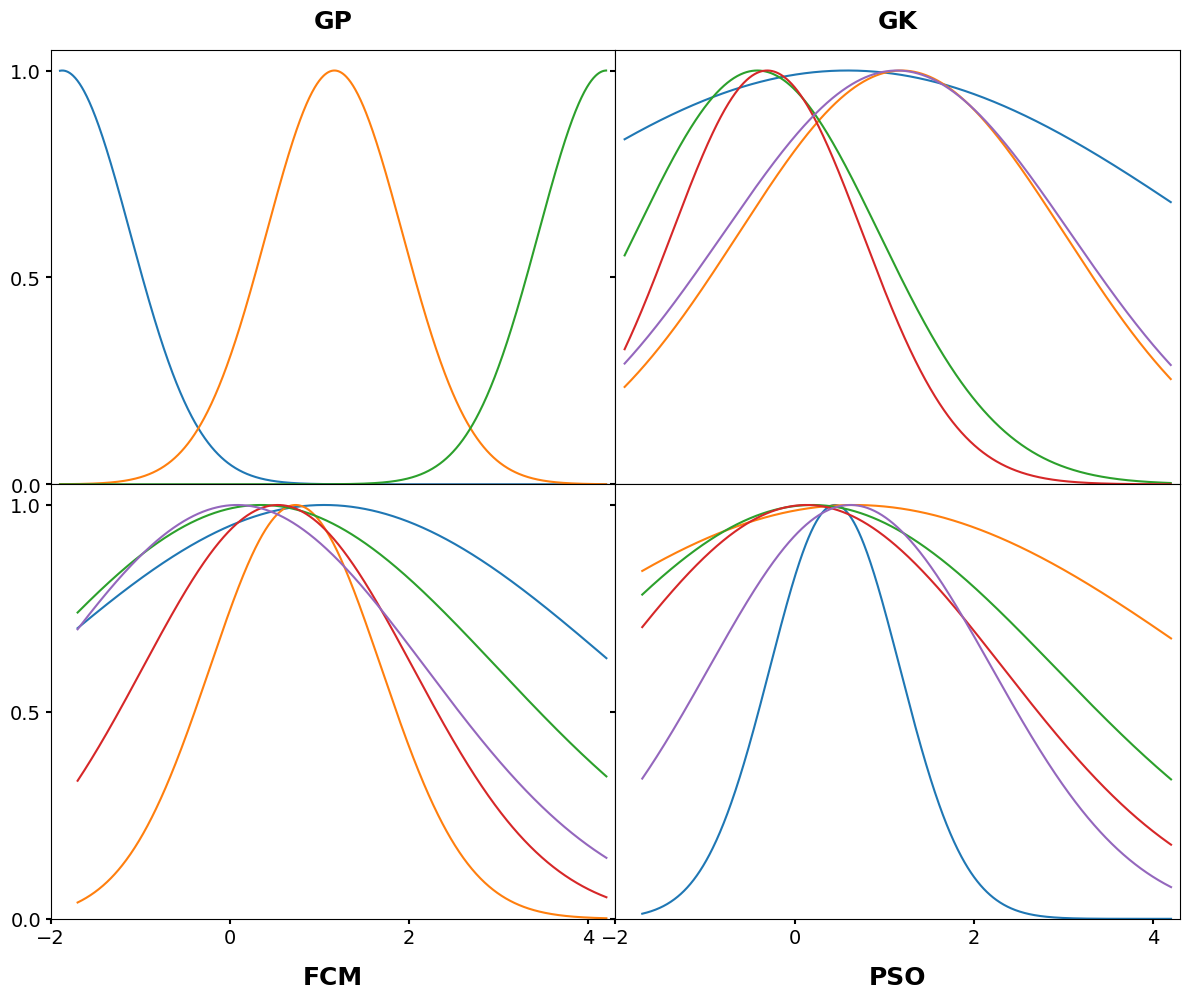}
    \caption{\small Fuzzy sets created for \(\Delta P\) using grid partitioning GP, Gustaffson-Kessel clustering GK, Fuzzy C-Means FCM and particle swarm optimization PSO.}
    \label{fig:distinguish}
\end{figure}
With respect to performance, the clustering model achieved an MAE of 0.49, corresponding to a MAPE of approximately 8.2\%. While this is significantly better than the pre-optimization grid-partitioning model, the optimized grid-partitioning model outperformed the clustering model, achieving a superior MAPE of 7.0\%.

For structural interpretability, both models had the same initial rule count, but simplification reduced rules to 15 in IA versus 18 in GRABS. The grid-partitioning method needed significantly fewer parameters than clustering-based methods, indicating better structural interpretability at both rule base and fuzzy partition levels per Gacto's complexity quadrants. Finally, to assess semantic interpretability, similarity metric $d(A,B)$ was used to quantify distinguishability. The grid partitioning model achieved a mean similarity of 0.115, significantly below the recommended threshold of 0.45 \cite{jin2000fuzzy}, which was suggested to achieve a good balance between system performance and interpretability. In contrast, the GK clustering-based model exhibited a higher mean similarity of 2.610, suggesting excessive overlap between membership functions, likely due to its sensitivity to data skewness. As shown in Figure \ref{fig:distinguish}, the grid model preserved more distinct and interpretable fuzzy sets, while the clustering models show considerable overlap. The two additional clustering methods, FCM and PSO, also underperformed in distinguishability compared to GK clustering.

\section{Conclusion and Future research}
Our study reveals that combining expert-guided grid partitioning with data-driven methods achieves performance comparable to clustering-based approaches, while offering superior semantic interpretability in forward osmosis desalination modeling. The results showed advantages in preserving the distinguishability of the membership functions, particularly in data sets with skewed distributions, where clustering methods encountered challenges. These findings suggest that integrating domain expertise into model design could help achieve a balance between accuracy and interpretability in fuzzy systems. This approach indicates that it is possible to maintain predictive performance while ensuring interpretability, which could offer a useful framework for water treatment applications, where both performance and interpretability are important. Future work would focus on refining the grid partitioning process under expert constraints and exploring its potential for reverse osmosis and ultra-filtration systems.
\vspace{-2mm}

\section*{Acknowledgment}
\small
This publication is part of the project Innovation Lab for Utilities on Sustainable Technology and Renewable Energy project (ILUSTRE), of the research programme LTP ROBUST which is partly financed by the Dutch Research Council (NWO).
\vspace{-6mm}
\section*{}
\bibliographystyle{IEEEtran}
\bibliography{ref}

\end{document}